\documentclass[conference]{IEEEtran}
\IEEEoverridecommandlockouts
\usepackage{cite}
\usepackage{amsmath,amssymb,amsfonts}
\usepackage{algorithmic}
\usepackage{graphicx}
\usepackage{textcomp}

\usepackage{booktabs}
\usepackage{multirow}
\usepackage{hyperref}
\usepackage{stfloats}
\usepackage{amsmath} 
\usepackage{url}
\usepackage[table,xcdraw]{xcolor}
\def\BibTeX{{\rm B\kern-.05em{\sc i\kern-.025em b}\kern-.08em
    T\kern-.1667em\lower.7ex\hbox{E}\kern-.125emX}}
\begin{document}

\title{Efficient Knowledge Transfer in Multi-Task Learning through Task-Adaptive Low-Rank Representation
\thanks{${\dagger}$: Corresponding author

This work was supported in part by National Nature Science Foundation of China (No. 62172036) and National Science and Technology Major Project (2022ZD0116305)}
}
\author{
    \IEEEauthorblockN{
        \textit{Xiao Zhang},
        \textit{Kangsheng Wang},
        \textit{Tianyu Hu}\textsuperscript{${\dagger}$}, 
        \textit{Huimin Ma}\textsuperscript{${\dagger}$}
    } \\
    \vspace{-10pt}
\IEEEauthorblockA{
        \textit{School of Computer and Communication Engineering} \\
        \textit{University of Science and Technology Beijing}\\
        Beijing, China \\
        \\
        \vspace{-35pt}
    }
    }

\maketitle

\begin{abstract}
   Pre-trained language models (PLMs) demonstrate remarkable intelligence but struggle with emerging tasks unseen during training in real-world applications. Training separate models for each new task is usually impractical. Multi-task learning (MTL) addresses this challenge by transferring shared knowledge from source tasks to target tasks. As an dominant parameter-efficient fine-tuning method, prompt tuning (PT) enhances MTL by introducing an adaptable vector that captures task-specific knowledge, which acts as a prefix to the original prompt that preserves shared knowledge, while keeping PLM parameters frozen. However, PT struggles to effectively capture the heterogeneity of task-specific knowledge due to its limited representational capacity. To address this challenge, we propose Task-Adaptive Low-Rank Representation (TA-LoRA), an MTL method built on PT, employing the low-rank representation to model task heterogeneity and a fast-slow weights mechanism where the slow weight encodes shared knowledge, while the fast weight captures task-specific nuances, avoiding the mixing of shared and task-specific knowledge, caused by training low-rank representations from scratch. Moreover, a zero-initialized attention mechanism is introduced to minimize the disruption of immature low-rank components on original prompts during warm-up epochs. Experiments on 16 tasks demonstrate that TA-LoRA achieves state-of-the-art performance in full-data and few-shot settings while maintaining superior parameter efficiency. \footnotemark[1]
\end{abstract}
\footnotetext[1]{\url{https://anonymous.4open.science/r/TA-LoRA-EB27}}

\section{Introduction}

Pre-trained language models (PLMs) exhibit human-like intelligence by capturing general knowledge from large-scale corpora \cite{brown2020language}. However, their applications remain largely confined to applications such as chatbots due to the complexity of real-world implementation, which often requires handling multiple downstream tasks, such as writing and reasoning, alongside emerging tasks unseen during training \cite{WANG202351, zhang2025enhancing, hu2025autonomous, wang2024credes}. The approach employed by mainstream PLM providers—training new models for new tasks such as data analysis and numerical prediction—is both costly and somewhat impractical \cite{openai2024o1, openai2024canvas}. Moreover, it does not effectively leverage cross-task knowledge. To address this, multi-task learning (MTL) offers a promissing solution by utilizing shared knowledge from source tasks (those available during training) to enhance generalization to target tasks (those involving unseen data).

Prompt tuning (PT) \cite{lester-etal-2021-power}, a parameter-efficient fine-tuning (PEFT) method, reduces the cost of adapting PLMs to downstream tasks by introducing a learnable continuous prompt vector as a prefix to the original prompt, while keeping the pre-trained parameters frozen. The adaptable prompt encodes task-specific knowledge, whereas the original prompt retains shared knowledge, aligning naturally with the objectives of MTL. Recent studies \cite{vu-etal-2022-spot, wang2023multitask} have explored fine-tuning adaptable vectors on multi-source tasks to generalize to target tasks, allowing PLMs to benefit from the knowledge transfer. However, even when learnable, adaptable prompts often struggle to capture the heterogeneity within source task distributions, possibly due to the limitations in expressive capacity imposed by their form and length \cite{chen2024efficient, wang2024universality, petrov2023prompting}. Heterogeneity refers to the differences among tasks, highlighting the unique feature distribution of each task. Ineffectively capturing heterogeneity can obscure the distinction between shared and task-specific knowledge, hinder the abstraction of shared knowledge, and limit the generalization to unseen tasks.

To address these challenges, we introduce Task-Adaptive Low-Rank Representation (TA-LoRA), a novel MTL method built on PT, effectively separating shared knowledge from heterogeneity by leveraging the consistent direction of adaptable vectors in the parameter space. We employ low-rank representations, expressed as the product of low-rank matrices, to approximate heterogeneity, as the finite degrees of freedom inherent in low-rank structures can effectively capture the primary directions of variation in parameter updates \cite{he2022towards}. These directions represent the evolving trends of task-specific feature distributions within the model's parameter space.

However, the low-rank matrices trained from scratch above result in the entanglement of shared and task-specific knowledge. To address this, we propose a fast-slow weights mechanism where the slow and fast weights represent shared and task-specific, respectively. Furthermore, we mitigate the disruption of low-rank representations to the original prompt using a zero-initialized attention mechanism \cite{zhang2024llama} to ensure stability during warm-up epochs. This design allows the low-rank representation weights to gradually amplify via a learnable gating factor. By efficiently decoupling shared and task-specific knowledge in multi-source tasks, TA-LoRA effectively generalizes shared knowledge to target tasks.

Extensive experiments were conducted on 16 NLP tasks, following the setup of \cite{sun-etal-2023-multitask}. The results on unseen data and unseen tasks demonstrate that TA-LoRA outperforms strong baselines while maintaining superior parameter efficiency. In specific benchmark scenarios, the proposed method even outperforms full fine-tuning. Notably, TA-LoRA remains effective in few-shot settings, achieving strong performance with only 32-shot on target tasks. Our main contributions are as follows.
\begin{figure*}
   \centering
   \vspace{-20pt}
   \includegraphics[width=\linewidth]{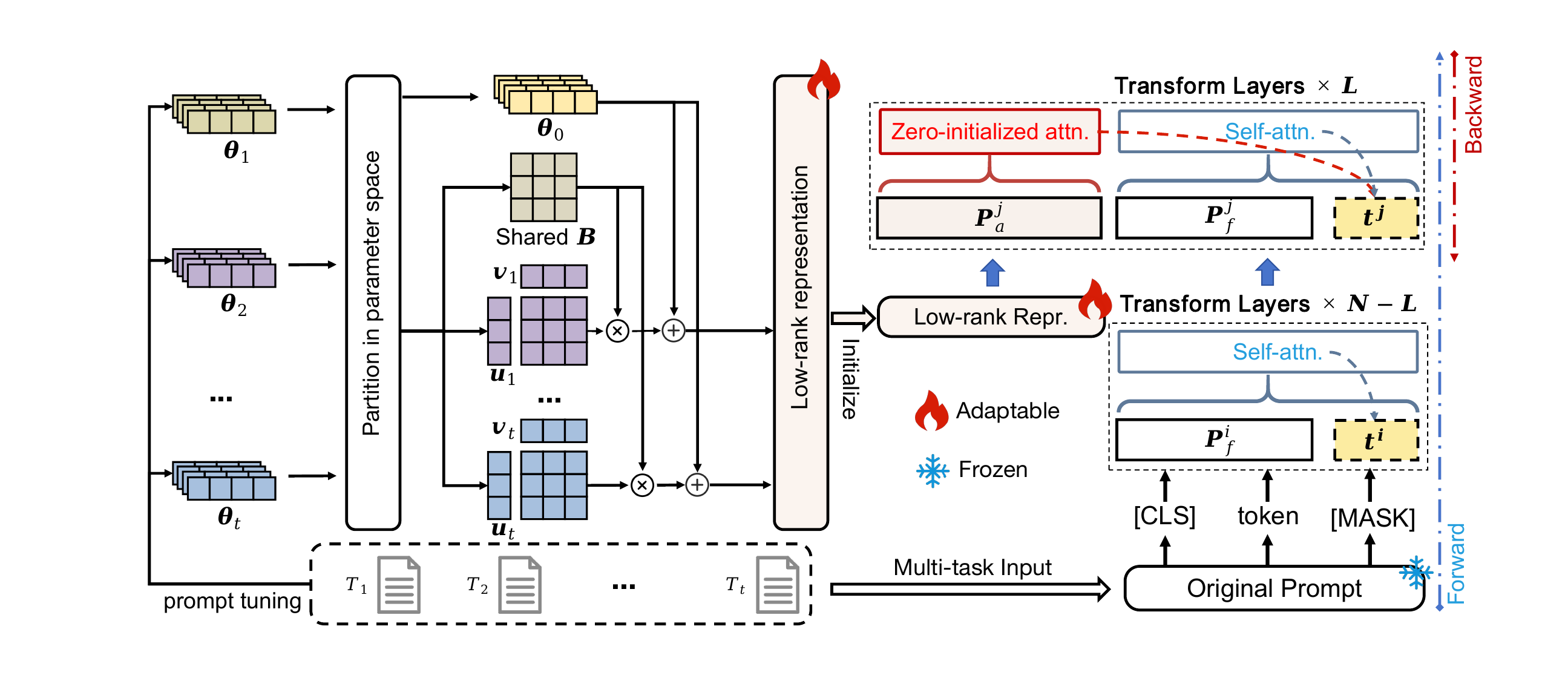}
   \vspace{-20pt}
   \caption{The framework of TA-LoRA serves as a plugin in the final $L$ layers of the PLM. ``Repr.'' and ``attn.'' stand for representation and attention, respectively.}\label{fig:framework}
   \vspace{-15pt}
   \end{figure*}

\begin{itemize} 
   \item We propose TA-LoRA, the first multi-task prompt tuning framework to leverage the low-rank representations for capturing task heterogeneity, enabling refined task-specific learning and enhanced shared knowledge abstraction for target task generalization.
   \item We propose a fast-slow weights mechanism to avoid mixing shared and task-specific knowledge during training from scratch, with slow weights encoding shared knowledge and fast weights capturing task-specific knowledge.
   \item Extensive experiments on 16 NLP tasks demonstrate that TA-LoRA outperforms strong baselines, achieving state-of-the-art performance while maintaining superior parameter efficiency.
\end{itemize}

\section{Background}
\textbf{Parameter-efficient fine-tuning.} Parameter-efficient fine-tuning is proposed to improve the efficiency of full fine-tuning by adjusting only a few parameters, while freezing the PLMs' parameters to retain pre-trained knowledge \cite{10687936}. BitFit \cite{ben-zaken-etal-2022-bitfit} adjusts only bias terms to enhance update efficiency, while LST \cite{LST} trains a small side network with shortcut connections from the backbone's intermediate activations. Methods like Adapter \cite{houlsby2019parameter, karimi-mahabadi-etal-2021-parameter} and its variants \cite{pfeiffer2020adapterfusion} inject tunable parameters into PLMs for downstream tasks. Among these, LoRA, which employs low-rank matrices to approximate model parameters, has gained significant attention. Another approach, Prompt tuning \cite{lester-etal-2021-power, liu-etal-2022-p}, introduces learnable prompt vectors, with Vanilla prompt tuning \cite{lester-etal-2021-power} adding prompt vectors before key-value matrices and P-Tuning \cite{liu-etal-2022-p} modifying the embedding matrix. TA-LORA improves by addressing the sensitivity to the initialization of adaptable prompts by applying a zero-initialized attention mechanism.  

\noindent\textbf{Multi-task prompt tuning.} Prompt tuning is particularly suitable for MTL compared to other PEFT methods because the adaptable prompt naturally disentangles shared and task-specific knowledge, with the position of the adaptable prompt relative to the original prompt in the model reflecting these two types of knowledge \cite{10688296, 10688334}. This makes it inherently capable of representing heterogeneity across multiple source tasks, a property proven effective in NLP. Common methods like SPoT \cite{vu-etal-2022-spot} use similarity metrics to select a prefix prompt for the current task, while ATTEMPT \cite{asai-etal-2022-attempt} interpolates prompts from large-scale source tasks with a newly initialized target task prompt, using instance attention from a lightweight sub-network trained on multiple target tasks. MPT \cite{wang2023multitask} learns a single transferable prompt by extracting knowledge from multiple task-specific source prompts. The proposed approach is the first to employ low-rank representations to capture task heterogeneity within PT, while avoiding the entanglement of shared and task-specific knowledge during training low-rank matrices from scratch, through a fast-slow weights mechanism.

\section{Methodology}

\subsection{Problem Formulation} 
Given a set of multi-source tasks $\boldsymbol{T} = \{\boldsymbol{T}_1, ..., \boldsymbol{T}_t\}$ exhibiting heterogeneity, to enable cross-task knowledge transfer in $\boldsymbol{T}$, we insert the adaptable continuous prompt vectors $\{\boldsymbol{\theta}_1, ..., \boldsymbol{\theta}_t\}$ initialized by PT on each task, as prefixes to original prompts. We propose TA-LoRA based on low-rank representation to capture the heterogeneity across $\boldsymbol{T}$, as illustrated in Figure~\ref{fig:framework}. We decompose the adaptable continuous prompt vectors in the parameter space into $\boldsymbol{\theta}_0$ (shared knowledge) and parameters capturing the heterogeneity of multi-source data, which enables the separation of shared and task-specific knowledge. The task-specific parameters are approximated via low-rank representation using a low-rank structure $\boldsymbol{B}\boldsymbol{A}$. 

Moreover, we introduce a fast-slow weights mechanism, splitting the low-rank representation matrices into a shared matrix $\boldsymbol{B}$ and task-specific matrices $\{\boldsymbol{A}_i\}_{i=1}^t$. To reduce computational complexity, we represent each $\boldsymbol{A}_i$ as a rank-1 matrix obtained from the outer product of two vectors $\boldsymbol{u}_i$ and $\boldsymbol{v}_i$. After multiplying the shared and task-specific low-rank matrices, they are added to $\boldsymbol{\theta}_0$, resulting in a low-rank representation. At the $l$-th layer, the adaptable prompt $\boldsymbol{P}_a^l$ serves as a prefix to the original prompt embedding $\boldsymbol{P}_f^l$. The prediction for the masked token is denoted as $\boldsymbol{t}^l$. Finally, TA-LoRA separately calculates the fine-tuned adaptable vector attention score using a zero-initialized attention mechanism to mitigate sensitivity to initialization.


\subsection{Low-rank Heterogeneity Representation}
We treat the adaptable vectors $\boldsymbol{\theta} = \{\boldsymbol{\theta}_1, ..., \boldsymbol{\theta}_t\}$ across tasks as a means to fine-tune PLMs and base models for capturing heterogeneous knowledge. Given $t$ base models, we average their parameters in the parameter space, decomposing the parameter set of each base model into two components: $\boldsymbol{\theta}_0 = \frac{1}{t} \sum_{i=1}^{t} \boldsymbol{\theta}_i$, which represents global task knowledge, and $\boldsymbol{\theta}_i - \boldsymbol{\theta}_0$, which captures task-specific knowledge, as shown in \eqref{eq:theta1}. This distinction is made because global task knowledge is shared across all tasks and can be represented as a consistent direction in the parameter space.
\vspace{-5pt}
\begin{equation}
   \boldsymbol{\theta} = \bigoplus_{i=1}^t \left[ \boldsymbol{\theta}_0 + \frac{1}{t} \sum_{j=1}^{t} \left(\boldsymbol{\theta}_i - \boldsymbol{\theta}_j\right) \right] \label{eq:theta1}
\end{equation}
\vspace{-5pt}
where $\bigoplus$ represents the concatenation operation along axis 0.

Due to the limited representational capacity of parameters of PT, capturing the differences between multi-source heterogeneous tasks becomes challenging. To address this, we introduce a low-rank representation in TA-LoRA to approximate the task-specific knowledge captured by these differences, as shown in \eqref{eq:lora0}. Therefore, \eqref{eq:theta1} is reformulated as \eqref{eq:theta2}. The low-rank representation approximates the task-specific knowledge, which is then combined with the shared knowledge $\boldsymbol{\theta}_0$ by a broadcasting mechanism to fine-tune PLMs.
\vspace{-5pt}
\begin{equation}
   \frac{1}{t}\sum_{j=1}^t(\boldsymbol{\theta}_i-\boldsymbol{\theta}_j) \approx \boldsymbol{B}_i\boldsymbol{A}_i \label{eq:lora0}
\end{equation}
\vspace{-12pt}
\begin{equation}
   \boldsymbol{\theta} = \boldsymbol{\theta}_0 + s \bigoplus_{i=1}^t \boldsymbol{B}_i\boldsymbol{A}_i \label{eq:theta2}
\end{equation}

where $\boldsymbol{B} \sim \mathcal{N}(0, \sigma^2)$ and $\boldsymbol{A} \sim \mathcal{N}(0, \sigma^2)$ represent low-rank matrices; $s$ is a scaling factor that controls the magnitude of the task-specific knowledge. 

When representing the heterogeneity of $\boldsymbol{T}$ using a low-rank approximation, we drew inspiration from \cite{chen2024efficient}: initially, the similarity between base models is nearly zero, but increases significantly as training progresses, particularly in lower-level layers. To demonstrate the suitability of Qwen (PLM with 28 decoder layers) \cite{qwen2.5} used in the proposed framework, we calculated the similarity of the 1st, 14th, and 28th layers, as shown in Appendix A \footnotemark[1], verifying the above hypothesis, which led us to deconstruct the low-rank representation in the later $L$ layers of Qwen. The high similarity in the first $28-L$ layers suggests that they capture shared features, while the later $L$ layers focus on task-specific features.

\subsection{Slow-Fast Weights Mechanism}
\begin{figure}
\centering
\vspace{-20pt}
\includegraphics[width=\columnwidth]{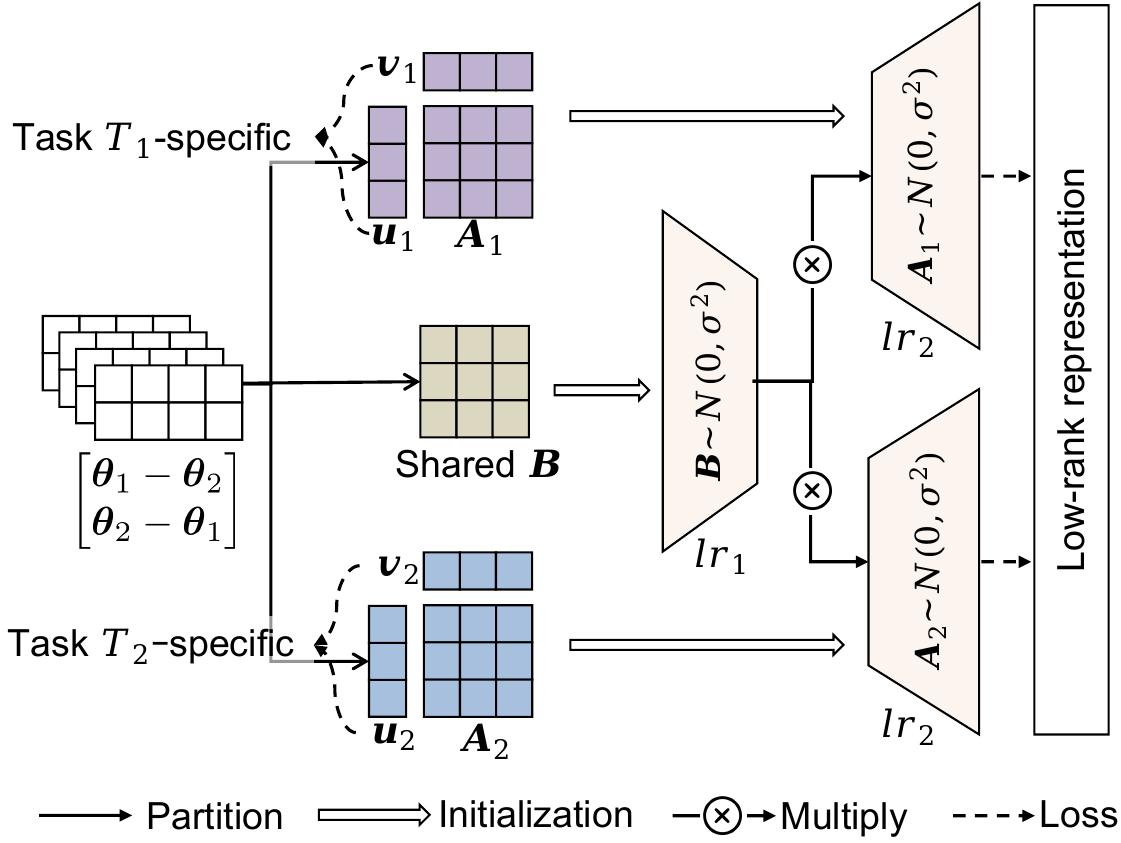}
\vspace{-20pt}
\caption{Decomposition of the slow weight $\boldsymbol{B}$ and the fast weight $\boldsymbol{A}_i$ for $\boldsymbol{T}_i$, with further decomposition of fast weight $\boldsymbol{A}_i$ into $\boldsymbol{u}_i$ and $\boldsymbol{v}_i$. ``lr'' stands for the learning rate.} \label{fig:decom}
\vspace{-18pt}
\end{figure}

Low-rank representation presents a challenge: capturing heterogeneity through low-rank approximation often causes the low-rank matrices, trained from scratch, to learn representations that entangle both cross-task knowledge and task-specific knowledge \cite{wang2023multitask}. This process is not affected by the previous parameter decomposition. To address this and reduce the computational complexity, we propose a fast-slow weights mechanism where $\boldsymbol{B}$ is treated as the slow weight shared across tasks, while $\boldsymbol{A}=\{\boldsymbol{A}_i\}_{i=1}^t$ serves as the fast weight to encode task-specific knowledge in the low-rank subspace for each task, as illustrated in Figure~\ref{fig:decom}. $\boldsymbol{B}$ and $\boldsymbol{A}$ are assigned with different learning rates to balance the varying gradient scales. To reduce the computational complexity introduced by $\boldsymbol{A}$ and to achieve task-specific knowledge specialization, we propose a compact representation method that decomposes the fast weight $\boldsymbol{A}_i$ for task $\boldsymbol{T}_i$ into two components: $\boldsymbol{u}_i$ and $\boldsymbol{v}_i$. The outer product of the task-specific vectors $\boldsymbol{u}_i$ and $\boldsymbol{v}_i$ forms the fast weight $\boldsymbol{A}_i = \boldsymbol{u}_i \otimes \boldsymbol{v}_i$, yielding a rank-1 matrix. Accordingly, \eqref{eq:lora0} is reformulated as follows.
\vspace{-5pt}
\begin{equation}
   \frac{1}{t}\sum_{j = 1}^t(\boldsymbol{\theta}_i - \boldsymbol{\theta}_j) \approx \boldsymbol{B}(\boldsymbol{A}_i) = \boldsymbol{B}(\boldsymbol{u}_i \otimes \boldsymbol{v}_i) \label{eq:lora1}
\end{equation}
\vspace{-10pt}

This approach effectively captures the core features of each task while minimizing the number of degrees of freedom. In MTL, the rank-1 matrix constraint regulates the model's capacity, encouraging it to prioritize the extraction of essential task features while avoiding overfitting on excessive detail or noise in the training data \cite{10688072,wang2025csceboostingllmreasoning}.

\subsection{Zero-Initialized Attention}
Low-rank representation poses an additional challenge: when randomly initialized, the low-rank matrices, being part of the prompt, may interfere with the original tokens, potentially leading to unreliable adapter prompts that disrupt attention calculation. Thus, we introduce a zero-initialized attention mechanism designed to minimize the disruption of unreliable low-rank matrices on the original prompt tokens during warm-up epochs. Specifically, we adapt the self-attention mechanism in the final $L$ layers to its variant of zero-initialized attention for the adaptable vector. The embedding $[\boldsymbol{P}_a^l, \boldsymbol{P}_f^l]$ are processed through multiple linear projection layers to generate the queries, keys, and values, as outlined below.
\begin{equation}
\boldsymbol{Q}^l = \boldsymbol{W}_q^l \cdot t^l \label{eq:q}
\end{equation}
\vspace{-12pt}
\begin{equation}
\boldsymbol{K}^l = \boldsymbol{W}_k^l \cdot [\boldsymbol{P}_a^l, \boldsymbol{P}_f^l] \label{eq:k}
\end{equation}
\vspace{-12pt}
\begin{equation}
\boldsymbol{V}^l = \boldsymbol{W}_v^l \cdot [\boldsymbol{P}_a^l, \boldsymbol{P}_f^l] \label{eq:v}
\end{equation}
where $\boldsymbol{W}_q^l$, $\boldsymbol{W}_k^l$, and $\boldsymbol{W}_v^l$ are the linear projection matrices for queries, keys, and values, respectively. The attention score is calculated as follows.
\begin{equation}
\boldsymbol{S}^l = \text{softmax}\bigr[\frac{\boldsymbol{Q}^l(\boldsymbol{K}^l)^T}{\sqrt{H}}\bigr]=[\boldsymbol{S}_a^l, \boldsymbol{S}_f^l] \label{eq:att}
\end{equation}

where $H$ denotes the feature dimension of PLMs; $\boldsymbol{S}_a^l$ and $\boldsymbol{S}_f^l$ are the attention scores for the adaptable and the original prompt tokens, respectively. At this stage, to ensure that the original tokens are not affected by unreliable low-rank matrices during the warm-up epochs, we set their attention scores to zero and gradually increase them as training progresses. We introduce a learnable gating factor, denoted as $g^l$ in \eqref{eq:gate}, which controls the attention scores of adaptable vectors. To prevent mutual interference between adaptable and original tokens, the $\text{softmax}(\cdot)$ function is applied independently to respective attention components $\boldsymbol{S}_a^l$ and $\boldsymbol{S}_f^l$.

\begin{equation}
\boldsymbol{S}_a^l \gets \text{tanh}(g^l) \cdot \boldsymbol{S}_a^l \label{eq:gate}
\end{equation}

\subsection{Source Optimization and Target Generalization} 
The optimization process for the source task consists of two steps. First, we train a single-source adapter prompt for each task, which serves as our base model. The base model is represented using \eqref{eq:theta2}, and the attention scores are calculated based on \eqref{eq:att}. Building on this, we optimize the proposed TA-LoRA framework by incorporating inter-task parameter orthogonalization as a regularization term into the loss function of the PLM, as detailed in \eqref{eq:loss}.
\begin{equation}
\mathcal{L} = \mathbb{E}_{(x, y) \in \boldsymbol{T}}\bigg[L_\text{PLM} + \lambda \sum_{i=1}^t\sum_{j\neq i} \|\boldsymbol{A}_i^\top\boldsymbol{A}_j-\boldsymbol{I}\|_2^2 \bigg] \label{eq:loss}
\end{equation}
Here, $L_\text{PLM}$ represents the pre-training loss \cite{qwen2.5} of the PLM, while the second term serves as a regularization component to enforce orthogonality among the task-specific low-rank matrices. The $\boldsymbol{A}_i^\top\boldsymbol{A}_j$ constraint is applied specifically to the final $L$ layers of the PLM decoder. The regularization coefficient $\lambda$ determines the relative weight of this term. Additionally, $\boldsymbol{B}$ and $\boldsymbol{A}$ are optimized with distinct learning rates to ensure balanced and effective training.

To achieve effective generalization for target tasks in a few-shot setting, we initialize a task-specific rank-1 $\boldsymbol{A}_i=(\boldsymbol{u_i}\otimes \boldsymbol{v_i})$ for each target task, and optimize it through $L_\text{PLM}$. Furthermore, we posit that if multiple target tasks are involved, the base model parameters for these tasks can be updated using the same approach as for the source tasks.

\section{Experiments}
\label{sec:results}
We evaluated TA-LoRA on a diverse set of NLP datasets intentionally selected to be independent of widely used benchmarks. This careful selection minimizes the risk of data leakage from the large-scale training corpora of PLMs, ensuring a fair assessment of the proposed method. Experimental results show that TA-LoRA consistently outperforms strong baselines in both full-data (Table~\ref{tab:full}) and few-shot (Table~\ref{tab:few}) settings, while maintaining superior performance efficiency.

\subsection{Experimental Setup}
\noindent\textbf{Datasets.} We curated and preprocessed 16 NLP task datasets from \cite{sun-etal-2023-multitask}, designating AFQMC, Amazon, THUCNews, BQ, CMNLI, CMRC-2018, SanWen, and COTE-MFW as source tasks, and ChnSent, TNews, OCNLI, LCQMC, DRCD, C3, COTE-BD, and FinRE as target tasks. Detailed descriptions of these datasets are provided in Appendix~B\footnotemark[1]. To evaluate TA-LoRA, we used two evaluation strategies: \verb|Unseen Data| and \verb|Unseen Task|. The former is a test set from the same distribution as the training data, evaluating the model's ability to capture heterogeneity, while the latter includes out-of-distribution data, assessing the generalization of shared knowledge to unknown tasks.

\noindent\textbf{Backbone.}
We use the PLM of Qwen2.5 \cite{qwen2.5} with 7B parameters and employ an adaptable prompt of length 20 and introduce our low-rank representation in the last 14 layers.

\noindent\textbf{Baselines.} We evaluate TA-LoRA against the following baselines: 
(1) Full fine-tuning (FT), where the entire PLM is fine-tuned;
(2) Adapter \cite{houlsby2019parameter}, where a tunable adapter is applied;
(3) BitFit \cite{ben-zaken-etal-2022-bitfit}, where only the bias parameters are fine-tuned;
(4) Vanilla prompt tuning (PT) \cite{lester-etal-2021-power}, where single-source prompts are fine-tuned;
(5) SPoT \cite{vu-etal-2022-spot}, where single-source adaptable prompts are applied;
(6) ATTEMPT \cite{asai-etal-2022-attempt}, where multi-source adaptable prompts are used;
(7) MPT \cite{wang2023multitask}, where multi-source adaptable prompts decomposition is employed.

\begin{table*}[t!]
\vspace{-20pt}
   \caption{
      Performance comparison of TA-LoRA with baselines on target tasks.
   } \label{tab:full}
   \vspace{-5pt}
   \setlength{\tabcolsep}{6.5pt}
   \centering
   \begin{tabular}{lcccccccccc}
   \toprule
   \multicolumn{11}{c}{\textbf{Unseen Data}}                                                                                                              \\ \midrule
   \textbf{Method} & \textbf{\begin{tabular}[c]{@{}c@{}}param/\\ task\end{tabular}} & \text{\begin{tabular}[c]{@{}c@{}}\textbf{AFQMC} \\ Acc.\end{tabular}}   & \text{\begin{tabular}[c]{@{}c@{}}\textbf{Amazon}\\ Acc.\end{tabular}} & \text{\begin{tabular}[c]{@{}c@{}}\textbf{THUCNews}\\ Acc.\end{tabular}} & \text{\begin{tabular}[c]{@{}c@{}}\textbf{BQ}\\ Acc.\end{tabular}}    & \text{\begin{tabular}[c]{@{}c@{}}\textbf{CMNLI}\\ Acc.\end{tabular}} & \text{\begin{tabular}[c]{@{}c@{}}\textbf{CMRC-2018}\\ F1\end{tabular}} & \text{\begin{tabular}[c]{@{}c@{}}\textbf{SanWen}\\ F1\end{tabular}}  & \text{\begin{tabular}[c]{@{}c@{}}\textbf{COTE-MFW}\\ F1\end{tabular}} & \multicolumn{1}{l}{\cellcolor[HTML]{C0C0C0}\textbf{Avg.}} \\ \midrule
   FT              & 7B                                                                                     & $\textbf{92.8}$                                                                                    & 61.7                                                                                   & 92.9                                                                                     & $\textbf{79.6}$                                                                                  & 63.4                                                                                  & 79.6                                                                                    & $\textbf{95.3}$                                                                                  & 87.2                                                                                   & \cellcolor[HTML]{C0C0C0}81.6                              \\
   Adapter        & 26M                                                                                    & 91.9                                                                                    & 60.3                                                                                   & 92.7                                                                                     & 69.9                                                                                  & 70.3                                                                                  & 69.8                                                                                    & 90.3                                                                                  & 86.7                                                                                   & \cellcolor[HTML]{C0C0C0}79.0                                \\
   BitFit          & 800K                                                                                   & 88.7                                                                                    & 63.2                                                                                   & 89.2                                                                                     & 69.3                                                                                  & 67.2                                                                                  & 77.3                                                                                    & 89.4                                                                                  & 86.3                                                                                   & \cellcolor[HTML]{C0C0C0}78.8                              \\
   PT          & 1.1M                                                                                   & 83.2                                                                                    & 56.3                                                                                   & 76.8                                                                                     & 67.8                                                                                  & 59.8                                                                                  & 73.6                                                                                    & 84.8                                                                                  & 85.2                                                                                   & \cellcolor[HTML]{C0C0C0}73.4                              \\
   SPoT            & 110K                                                                                   & 91.6                                                                                    & 61.4                                                                                   & 87.2                                                                                     & 71.8                                                                                  & 69.7                                                                                  & 76.2                                                                                    & 89.4                                                                                  & 81.5                                                                                   & \cellcolor[HTML]{C0C0C0}78.6                              \\
   ATTEMPT         & 910K                                                                                   & 91.2                                                                                    & 58.9                                                                                   & \textbf{93.2}                                                                                     & 69.4                                                                                  & 67.3                                                                                  & 77.5                                                                                    & 88.6                                                                                  & 88.8                                                                                   & \cellcolor[HTML]{C0C0C0}79.3                              \\
   MPT             & 1M                                                                                     & 91.8                                                                                    & 63.7                                                                                   & 92.0                                                                                       & 71.7                                                                                  & 69.5                                                                                  & 78.3                                                                                    & 92.1                                                                                  & 84.8                                                                                   & \cellcolor[HTML]{C0C0C0}80.5                              \\
   Ours            & 1.3M                                                                                   & $\text{92.4}_{\pm0.8}$                                                                                & $\textbf{67.3}_{\pm1.2}$                                                                      & $\textbf{93.2}_{\pm0.6}$                                                                        & $\text{78.4}_{\pm0.9}$                                                                              & $\textbf{73.3}_{\pm0.4}$                                                                     & $\textbf{83.1}_{\pm1.1}$                                                                       & $\text{90.4}_{\pm0.2}$                                                                              & $\textbf{89.5}_{\pm0.7}$                                                                      & \cellcolor[HTML]{C0C0C0}\textbf{83.4}                     \\ \midrule
   \multicolumn{11}{c}{\textbf{Unseen Task}}                                                                                                                         \\ \midrule
   \textbf{Method} & \textbf{\begin{tabular}[c]{@{}c@{}}param/\\ task\end{tabular}} & \text{\begin{tabular}[c]{@{}c@{}}\textbf{ChnSent}\\ Acc.\end{tabular}} & \text{\begin{tabular}[c]{@{}c@{}}\textbf{TNews}\\ Acc.\end{tabular}}  & \text{\begin{tabular}[c]{@{}c@{}}\textbf{OCNLI}\\ Acc.\end{tabular}}    & \text{\begin{tabular}[c]{@{}c@{}}\textbf{LCQMC}\\ Acc.\end{tabular}} & \text{\begin{tabular}[c]{@{}c@{}}\textbf{DRCD}\\ F1\end{tabular}}    & \text{\begin{tabular}[c]{@{}c@{}}\textbf{C3}\\ Acc.\end{tabular}}      & \text{\begin{tabular}[c]{@{}c@{}}\textbf{COTE-BD}\\ F1\end{tabular}} & \text{\begin{tabular}[c]{@{}c@{}}\textbf{FinRE}\\ F1\end{tabular}}    & \cellcolor[HTML]{C0C0C0}\textbf{Avg.}                     \\ \midrule
   FT              & 7B                                                                                     & 89.5                                                                                    & 51.4                                                                                   & 53.4                                                                                     & 72.5                                                                                  & 82.2                                                                                  & 48.1                                                                                    & 89.1                                                                                  & 71.7                                                                                   & \cellcolor[HTML]{C0C0C0}69.7                              \\
   Adapter        & 26M                                                                                    & 89.2                                                                                    & 53.9                                                                                   & 51.2                                                                                     & 71.9                                                                                  & 81.5                                                                                  & 47.2                                                                                    & 88.3                                                                                  & 71.5                                                                                   & \cellcolor[HTML]{C0C0C0}69.3                              \\
   BitFit          & 800K                                                                                   & 88.7                                                                                    & 50.8                                                                                   & 52.7                                                                                     & 71.4                                                                                  & 82.9                                                                                  & 48.3                                                                                    & 87.2                                                                                  & 70.3                                                                                   & \cellcolor[HTML]{C0C0C0}69.0                                \\
   PT          & 1.1M                                                                                   & 82.3                                                                                    & 51.5                                                                                   & 59.8                                                                                     & 76.8                                                                                  & 74.8                                                                                  & 34.6                                                                                    & 84.4                                                                                  & 69.6                                                                                   & \cellcolor[HTML]{C0C0C0}66.7                                \\
   SPoT            & 110K                                                                                   & 89.9                                                                                    & 52.3                                                                                   & 51.4                                                                                     & 73.8                                                                                  & 83.9                                                                                  & 50.3                                                                                    & 91.2                                                                                  & 72.3                                                                                   & \cellcolor[HTML]{C0C0C0}70.6                              \\
   ATTEMPT         & 910K                                                                                   & 88.2                                                                                    & 53.6                                                                                   & 51.8                                                                                     & 74.4                                                                                  & 84.7                                                                                  & $\textbf{51.0}$                                                                                      & 90.7                                                                                  & 74.3                                                                                   & \cellcolor[HTML]{C0C0C0}71.0                                \\
   MPT             & 1M                                                                                     & 90.0                                                                                      & 54.3                                                                                   & 52.9                                                                                     & 75.2                                                                                  & $\textbf{85.1}$                                                                                  & 50.0                                                                                      & 90.3                                                                                  & 75.9                                                                                   & \cellcolor[HTML]{C0C0C0}71.7                              \\
   Ours            & 1.3M                                                                                   & $\textbf{91.7}_{\pm0.3}$                                                                       & $\textbf{59.7}_{\pm2.7}$                                                                      & $\textbf{68.1}_{\pm1.1}$                                                                        & $\textbf{80.9}_{\pm0.7}$                                                                     & $\text{82.6}_{\pm0.9}$                                                                              & $\text{42.0}_{\pm1.7}$                                                                                & $\textbf{93.2}_{\pm0.2}$                                                                     & $\textbf{76.3}_{\pm1.4}$                                                                      & \cellcolor[HTML]{C0C0C0}\textbf{74.3}                     \\ \bottomrule
   \end{tabular}
   \vspace{-10pt}
   \end{table*}

\begin{table*}[]
   \centering
   \setlength{\tabcolsep}{10.65pt}
   \renewcommand{\arraystretch}{}
   \caption{Few-shot performance comparison of TA-LoRA with baselines on setting of Unseen Task.}
   \vspace{-5pt}
   \begin{tabular}{llccccccccc}
   \toprule
                                     &                                   & \multicolumn{9}{c}{\textbf{Unseen Task}}                                                                                                                                      \\ \cmidrule(lr){3-11} 
   \multirow{-2}{*}{\textbf{$k\text{-shot}$}} & \multirow{-2}{*}{\textbf{Method}} & \textbf{ChnSent} & \textbf{TNews} & \textbf{OCNLI} & \textbf{LCQMC} & \textbf{DRCD} & \textbf{C3} & \textbf{COTE-BD} & \textbf{FinRE} & \cellcolor[HTML]{C0C0C0}\textbf{Avg.} \\ \midrule
                                     & PT                                & 83.9             & 51.9           & 60.3           & 78.8           & 75.4          & 34.0        & 86.1             & 70.0           & \cellcolor[HTML]{C0C0C0}67.6          \\
                                     & MPT                               & 90.5               & 55.2           & 51.7           & 76.6           & \textbf{88.7}         & \textbf{50.3}          & 92.3             & 77.2           & \cellcolor[HTML]{C0C0C0}72.8          \\
   \multirow{-3}{*}{16}              & Ours                              & \textbf{92.7}    & \textbf{61.3}  & \textbf{71.4}  & \textbf{86.7}  & 85.3          & 44.6          & \textbf{93.1}    & \textbf{78.8}  & \cellcolor[HTML]{C0C0C0}\textbf{76.7} \\ \midrule
                                     & PT                                & 82.3             & 54.3           & 60.5           & 81.7           & 76.1          & 38.6        & 87.6             & 71.8           & \cellcolor[HTML]{C0C0C0}69.1          \\
                                     & MPT                               & 92.2               & 58.5           & 54.4           & 79.4           & 88.2          & 55.2        & \textbf{94.5}             & 80.3             & \cellcolor[HTML]{C0C0C0}75.3          \\
   \multirow{-3}{*}{32}              & Ours                              & \textbf{94.0}    & \textbf{63.5}  & \textbf{72.9}  & \textbf{86.9}  & \textbf{88.5}          & \textbf{60.3}          & 93.4    & \textbf{80.4}  & \cellcolor[HTML]{C0C0C0}\textbf{80.0} \\ \midrule
                                     & PT                                & 84.7             & 57.3           & 62.8           & 83.5           & 78.8          & 40.6        & 89.9             & 74.6           & \cellcolor[HTML]{C0C0C0}71.5          \\
                                     & MPT                               & 92.6               & 59.3           & 57.2           & 80.2           & 89.7          & 56.3        & \textbf{94.6}             & 81.7           & \cellcolor[HTML]{C0C0C0}76.5          \\
   \multirow{-3}{*}{64}              & Ours                              & \textbf{96.3}    & \textbf{62.7}  & \textbf{73.4}  & \textbf{86.3}  & \textbf{90.4}          & \textbf{61.5}        & 93.7    & \textbf{85.7}  & \cellcolor[HTML]{C0C0C0}\textbf{81.3} \\ \bottomrule
   \end{tabular} \label{tab:few}
   \vspace{-10pt}
   \end{table*}

   \begin{table*}[]
      \centering
      \setlength{\tabcolsep}{6.45pt}
      \caption{Ablation results on F\&SW and zero-initialized attn. F\&SW indicates fast-slow weights mechanism.} \label{tab:ablation}
      \begin{tabular}{c|ccccc|ccccc}
      \toprule
                                         & \multicolumn{5}{c|}{\textbf{Unseen Data}}                                                                     & \multicolumn{5}{c}{\textbf{Unseen Task}}                                                                     \\ \cmidrule(lr){2-11}
      \multirow{-2}{*}{\textbf{Setting}} & \textbf{BQ}   & \textbf{CMNLI} & \textbf{CMRC-2018} & \textbf{SanWen} & \cellcolor[HTML]{C0C0C0}\textbf{Avg.} & \textbf{ChnSent} & \textbf{LCQMC} & \textbf{DRCD} & \textbf{COTE-BD} & \cellcolor[HTML]{C0C0C0}\textbf{Avg.} \\ \midrule
      Ours                               & \textbf{78.4} & \textbf{73.3}  & \textbf{83.1}      & \textbf{90.4}   & \cellcolor[HTML]{C0C0C0}\textbf{81.3} & \textbf{91.7}    &  80.9  & \textbf{82.6} & \textbf{93.2}    & \cellcolor[HTML]{C0C0C0}\textbf{87.1} \\
      Ours w/o F\&SW LR                 & 78.2          & 71.6           & 80.0               & 89.4            & \cellcolor[HTML]{C0C0C0}79.8            & 87.2             & \textbf{84.6}           & 79.3          & 91.4             & \cellcolor[HTML]{C0C0C0}85.6             \\
      Ours w/o F\&SW $\boldsymbol{B}$                 & 69.4          & 68.3           & 79.6               & 86.6            & \cellcolor[HTML]{C0C0C0}76.0               & 89.2             & 79.7           & 78.5          & 90.2             & \cellcolor[HTML]{C0C0C0}84.4          \\
      
      Ours w/o Zero-initialized attn.    & 74.7          & 68.8           & 81.3               & 88.1            & \cellcolor[HTML]{C0C0C0}78.2          & 91.6             & 80.3           & 81.9          & 89.0               & \cellcolor[HTML]{C0C0C0}85.7          \\ \bottomrule
      \end{tabular}
      \vspace{-10pt}
      \end{table*}

\subsection{Main Results}

\noindent\textbf{Full-data adaptation.}
Table~\ref{tab:full} presents the performance of different methods across 16 tasks under two settings: \verb|Unseen Data| and \verb|Unseen Task|. The results demonstrate the superior performance and high parameter efficiency of TA-LoRA. Compared to PT, TA-LoRA achieves great improvements in performance while maintaining the same task-specific parameter scale. Specifically, TA-LoRA outperforms PT by approximately 13.6\% on unseen data and 11.4\% on unseen tasks, underscoring the effectiveness of extracting shared and task-specific knowledge from multi-source tasks. While Adapters also show competitive performance, the proposed method consistently outperforms it and offers superior parameter efficiency. In the \verb|Unseen Data| setting, TA-LoRA achieves an average performance that surpasses FT. While its performance is marginally lower than FT on certain data points, TA-LoRA utilizes only 0.1857\% of FT's parameters. In the \verb|Unseen Task| setting, TA-LoRA consistently outperforms FT, further demonstrating the benefits of leveraging cross-task knowledge transfer. Moreover, TA-LoRA achieves state-of-the-art results, surpassing SPoT, ATTEMPT, and MPT with a comparable parameter scale.

\noindent\textbf{Few-shot adaptation.}
We conducted few-shot experiments ($k=16,32,64$) to evaluate the generalization ability of TA-LoRA to unseen tasks with limited training examples. Table~\ref{tab:few} presents the results of our method compared to other baselines. The findings demonstrate that TA-LoRA consistently outperforms PT and MPT across various $k\text{-shot}$ settings. Notably, in some datasets, TA-LoRA with 32-shot achieves performance comparable to the full dataset setting. These findings clearly indicate that TA-LoRA effectively leverages cross-task knowledge from source tasks to generalize to target tasks with only a few labeled samples.

\subsection{Ablation Studies}

\noindent\textbf{Fast-slow weights mechanism.}
To verify the effectiveness of the fast-slow weights mechanism, we compare TA-LoRA with a version of the proposed method without this mechanism, as illustrated in Table~\ref{tab:ablation}. Moreover, we treat the shared and task-specific matrices as having the same learning rate, and the corresponding results (2nd row of Table~\ref{tab:ablation}) indicate that the baseline (1st row of Table~\ref{tab:ablation}) achieves an average performance improvement of 1.8\% and 1.7\% under the two settings compared to the ablation results. Additionally, the 3rd row of Table~\ref{tab:ablation} shows that removing the shared $\boldsymbol{B}$ mechanism significantly reduces the performance of the proposed method. 

\noindent\textbf{Zero-initialized attention.}
We further examined the disruption of zero-initialized attention on TA-LoRA, as shown in the 4th row of Table~\ref{tab:ablation}. When the attention mechanism in TA-LoRA was replaced with the default multi-head self-attention, we observed a performance decline. These findings confirm that zero-initialized attention enhances the performance of TA-LoRA, aligning with the results reported by \cite{zhang2024llama}, which demonstrated that zero-initialized attention mitigates the sensitivity of adaptable prompts to initialization.

\section{Conclusion}
The proposed method aims to capture heterogeneity across source tasks by leveraging the low-rank representation, effectively disentangling shared and task-specific knowledge. To achieve this, we propose TA-LoRA, a novel MTL approach. Moreover, we propose a fast-slow weights mechanism, where the slow weight encodes shared knowledge, and the fast weight captures task-specific knowledge, reducing their entanglement during training. Task-specific knowledge is regularized through interactions with other tasks, preventing overfitting to details and noise in datasets and enhancing robustness, while shared knowledge acts as prior information to facilitate generalization to target tasks. To address initialization sensitivity, we introduce a zero-initialized attention with a gating mechanism that separates the attention computation for adaptable and original prompts, preventing early-stage prompts from interfering with the original ones. Experimental results demonstrate that TA-LoRA outperforms existing methods in both full-data and few-shot scenarios, achieving superior generalization to target tasks while maintaining parameter efficiency.


\bibliographystyle{IEEEbib}
\bibliography{icme2025references}

\clearpage

\appendices
\renewcommand\thesubsection{\thesection.\arabic{subsection}}
\section{Interlayer Similarity}

\label{app:A}
\begin{table*}[b]
   
   \setlength{\tabcolsep}{15pt}
   \caption{Overview of the datasets utilized for training, \texttt{Unseen Data}, and \texttt{Unseen Task}. Each dataset spans diverse tasks and domains, serving as a comprehensive benchmark for assessing the performance of TA-LoRA.}
   \centering
   \begin{tabular}{clcccl}
   \toprule
   \textbf{ID} & \textbf{Dataset}   & \textbf{Task Type}              & \textbf{Domain}         & \textbf{Size}  & \textbf{Source/Link} \\ \midrule
   1  & AFQMC     & Semantic Matching       & Financial       & 38K   & \href{https://aclanthology.org/2020.coling-main.419/}{Xu et al. (2020)}  \\
   2  & Amazon    & Sentiment Analysis      & Shopping Reviews & 4.1M  & \href{https://www.example.com}{Github}  \\
   3  & THUCNews  & Text Classification     & General          & 55K   & \href{https://www.example.com}{Github}  \\
   4  & BQ        & Semantic Matching       & Financial        & 110K  & \href{https://aclanthology.org/D18-1536/}{Chen et al. (2018)}\\
   5  & CMNLI     & Natural Language Inference & General      & 404K  & \href{https://aclanthology.org/2020.coling-main.419/}{Xu et al. (2020)} \\
   6  & CMRC-2018 & Reading Comprehension   & General          & 11.9K & \href{https://aclanthology.org/2020.coling-main.419/}{Xu et al. (2020)}  \\
   7  & SanWen    & Relation Extraction     & Literature       & 16K   & \href{https://aclanthology.org/2020.coling-main.419/}{Xu et al. (2020)}  \\
   8  & COTE-MFW  & Opinion Mining          & Shopping Reviews & 37K   & \href{https://proceedings.mlr.press/v95/li18d.html}{Li et al. (2018)}  \\
   9  & ChnSent   & Sentiment Analysis      & Financial        & 12K   & \href{https://github.com/aceimnorstuvwxz/toutiao-text-classfication-dataset}{Github}  \\
   10 & TNews     & Text Classification     & General          & 63K   & \href{https://github.com/aceimnorstuvwxz/toutiao-text-classfication-dataset}{Github}  \\
   11 & OCNLI     & Natural Language Inference & Biomedical    & 53K   & \href{https://aclanthology.org/2020.findings-emnlp.314/}{Hu et al. (2020)}  \\
   12 & LCQMC     & Question Matching       & General          & 250K  & \href{https://aclanthology.org/C18-1166/}{Liu et al. (2018)}  \\
   13 & DRCD      & Reading Comprehension   & General          & 10K   & \href{https://arxiv.org/abs/1806.00920}{Shao et al. (2018)}  \\
   14 & C3        & Multi-choice QA         & General          & 11K   & \href{https://direct.mit.edu/tacl/article/doi/10.1162/tacl_a_00305/43546/Investigating-Prior-Knowledge-for-Challenging}{Sun et al. (2018)}  \\
   15 & COTE-BD   & Aspect-based Sentiment Analysis & History & 8K    & \href{https://proceedings.mlr.press/v95/li18d.html}{Li et al. (2018)}  \\
   16 & FinRE     & Relation Extraction     & Financial        & 14K   & \href{https://aclanthology.org/P19-1430/}{Li et al. (2019)}  \\  \bottomrule
   \end{tabular} \label{tab:data}
\end{table*}

Inspired by the findings in \cite{chen2024efficient}, where the similarity between base models was observed to be nearly zero initially but increased significantly during training, particularly in the lower-level layers, we further analyzed this phenomenon in the context of the proposed TA-LoRA framework. Specifically, we evaluated the similarity between base models at different layers of Qwen (a PLM with 28 decoder layers) \cite{qwen2.5} to demonstrate its suitability within our framework.
\begin{figure}[!h]
   \centering
   \includegraphics[width=0.49\textwidth]{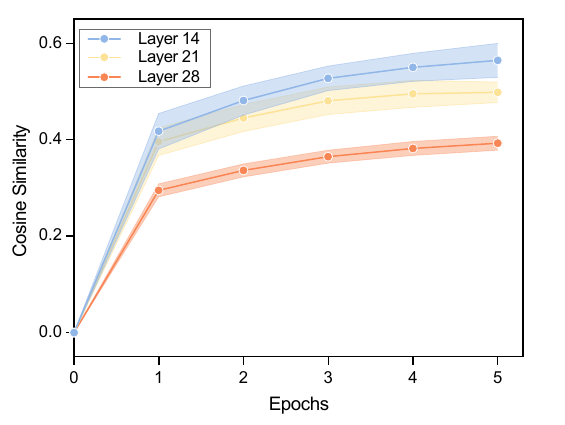}
   \caption{The similarity between base models in different layers obtained by PT on AFQMC.}
   \label{fig:sim}
\end{figure}

We calculated the similarity of the 1st, 14th, and 28th layers, as shown in Figure~\ref{fig:sim}. At the onset of training, the similarity between the base models is negligible, with values close to zero (e.g., -0.0015\% at the 14th layer, 0.0018\% at the 21st layer, and -0.0017\% at the 27th layer). For the analysis presented in Figure~\ref{fig:sim}, these similarities are approximated as zero for simplicity. Our results indicate that the similarity between base models in the 1st layer is almost negligible, reflecting the early layers' focus on general, shared knowledge across tasks. However, as training progresses, the similarity increases significantly in the 14th and 28th layers. This trend is particularly pronounced in the 28th layer, suggesting that these later layers prioritize task-specific features while still benefiting from shared representations learned earlier.

This observation aligns with the hierarchical nature of PLMs, where earlier layers capture more general features, and deeper layers specialize in task-specific nuances. It also underscores the effectiveness of the Qwen architecture in our framework, as the model's structure allows for a smooth transition from shared knowledge to task-specific representation, effectively capturing heterogeneity in the feature space. The increasing similarity in later layers reflects the ability of the low-rank representations in TA-LoRA to disentangle shared and task-specific knowledge, ensuring both effective cross-task knowledge transfer and specialization for individual tasks. This supports the rationale behind using Qwen in our proposed method and highlights its capability to address the challenges of multi-source task adaptation.

\section{Dataset Details}
\label{app:B}

This section provides detailed information about the datasets used for training, unseen data evaluation, and unseen task evaluation. The datasets span diverse tasks, domains, and sizes, offering a comprehensive benchmark for assessing the performance of TA-LoRA.

\subsection*{B.1 Training Data}
We utilize the training sets of eight source tasks for multi-task learning: AFQMC, Amazon, THUCNews, BQ, CMNLI, CMRC-2018, SanWen, and COTE-MFW. These datasets encompass various tasks such as semantic matching, sentiment analysis, reading comprehension, and text classification. The diversity of these tasks enables the model to learn shared representations across tasks while capturing task-specific nuances. Additionally, small-scale datasets are augmented using techniques like up-sampling and data enhancement methods, including synonym substitution and random addition or deletion. For large-scale datasets, down-sampling is applied to achieve balance.

\textbf{-\! Task Types:} These tasks are designed to evaluate different aspects of language understanding, such as recognizing semantic equivalence (e.g., AFQMC, BQ), opinion mining (e.g., COTE-MFW), and information extraction (e.g., CMRC-2018).

\textbf{-\! Domains:} The datasets cover a wide range of domains, including financial text, shopping reviews, and general literature, ensuring that the model is exposed to diverse linguistic styles and vocabularies.

\textbf{-\! Dataset Statistics:} As shown in Table~\ref{tab:data}, the sizes of the training sets vary significantly, ranging from 11.9K (CMRC-2018) to 4.1M (Amazon). This variability provides an opportunity to test the model's robustness across tasks of different scales.

\subsection*{B.2 Target Tasks of Unseen Data}
To evaluate the model's ability to generalize to unseen data within the same tasks, we use the validation sets of the eight source tasks (AFQMC, Amazon, THUCNews, BQ, CMNLI, CMRC-2018, SanWen, and COTE-MFW). These validation sets are excluded from the training process and serve as test sets for the unseen data evaluation strategy.

\textbf{-\! Purpose:} This evaluation assesses how well the model performs on held-out data from tasks it has already encountered during training. It focuses on the model's ability to generalize without overfitting to the training data.

\textbf{-\! Dataset Characteristics:} The validation sets are smaller in size compared to the training sets, making them suitable for evaluating the model's precision on limited data. For example, AFQMC's validation set consists of 38K samples, while COTE-MFW includes 37K samples.

\subsection*{B.3 Target Tasks of Unseen Tasks}
To assess the model's capability to generalize across tasks, we employ the training sets of eight downstream tasks: ChnSent, TNews, OCNLI, LCQMC, DRCD, C3, COTE-BD, and FinRE. These tasks are intentionally chosen to be distinct from the source tasks, covering new task types and domains.

\textbf{-\! Task Types:} The unseen tasks include sentiment analysis (e.g., ChnSent), natural language inference (e.g., OCNLI), and machine reading comprehension (e.g., DRCD, C3), which differ from the source tasks in both format and objective.

\textbf{-\! Domains:} These datasets span financial text (e.g., FinRE, ChnSent), biomedical text (e.g., OCNLI), and general knowledge (e.g., TNews, LCQMC), allowing the evaluation of domain transfer capabilities.

\textbf{-\! Dataset Statistics:} As shown in Table~\ref{tab:data}, these datasets vary in size from 8K (COTE-BD) to 250K (LCQMC), providing a challenging benchmark for cross-task generalization.

\subsection*{B.4 Analysis and Implications}
The dataset selection in Table~\ref{tab:data} ensures a balance between task and domain diversity, as well as variability in data size. This design serves several key purposes:

\textbf{-\! Diversity in Learning:} By exposing the model to tasks from different domains and linguistic structures, we ensure that TA-LoRA learns robust shared representations that generalize well across tasks.

\textbf{-\! Scalability Testing:} The wide range of dataset sizes allows us to test the scalability of TA-LoRA, ensuring that it performs consistently regardless of the amount of data available for a task.

\textbf{-\! Cross-Domain Generalization:} The inclusion of tasks from distinct domains (e.g., financial, biomedical, and general text) highlights the model's ability to adapt to unseen contexts, a crucial feature for practical applications.

The comprehensive evaluation using these datasets demonstrates the effectiveness of TA-LoRA in leveraging multi-task learning to achieve superior generalization, as detailed in Section~\ref{sec:results}.

\end{document}